\documentclass{article}%
\usepackage[semicolon]{natbib}
\usepackage{amsmath}
\usepackage{amsfonts}
\usepackage{amssymb}
\usepackage{xcolor}
\usepackage{graphicx}
\usepackage[ruled,linesnumbered]{algorithm2e}
\usepackage{enumitem}
\usepackage{hyperref}
\usepackage{tikz-cd}
\usepackage{setspace}%
\providecommand{\U}[1]{\protect\rule{.1in}{.1in}}
\makeatletter
\newcommand\setItemnumber[1]{\setcounter{enum\romannumeral\@enumdepth}{\numexpr#1-1\relax}}
\makeatother
\providecommand{\U}[1]{\protect\rule{.1in}{.1in}}
\SetKwComment{Comment}{/* }{ */}

\newtheorem{example}{Example}[section]
\newtheorem{problem}{Problem}[section]

\newtheorem{remark}{Remark}[section]
\newtheorem{conjecture}{Conjecture}[section]

\newtheorem{definition}{Definition}[section]
\definecolor{deepblue}{rgb}{0.1, 0.2, 0.7}
\hypersetup
{
colorlinks,
citecolor = deepblue,
filecolor = deepblue,
linkcolor = deepblue,
urlcolor = deepblue,
}

\usetikzlibrary{arrows,positioning,automata}
\begin{document}

\title{An entropic explanation of insistence on sameness in autism}
\author{{\small Przemysław Śliwiński\thanks{\href{https://orcid.org/0000-0003-3839-1580}{0000-0003-3839-1580}. E-mail: \href{mailto:przemyslaw.sliwinski@pwr.edu.pl}{\texttt{przemyslaw.sliwinski@pwr.edu.pl}}, Faculty of Information and Communication Technology, Wrocław University of Science and Technology, Wrocław, Poland.}}}
\date{{\small December 15, 2025}}
\maketitle

\begin{abstract}


\noindent\textbf{Purpose:} An information theory-based framework is proposed
in attempt to explain \emph{insistence on sameness in autism} as an instance
of a general behavior pattern in which an individual tries to reduce surprise and uncertainty. It offers a new definition of autism as \emph{an
impairment in which cognitive functions are restricted to discrimination,
memorization and prediction of tangible properties of the environment}.
\vspace*{1mm}

\noindent\textbf{Methods:} An analogy between \emph{insistence on sameness}
and \emph{constrained minimization} of the \emph{entropy metric} is observed
and examined for a set of assumptions that describe cognitive limitations of a
person with autism. The metric is given by the formula%
\[
D_{H}(R,M)=H(R|M)+H(M|R),
\]
where $R$ represents sequences of random stimuli, $M$ is a memory that stores
and retrieves them, and where $H(\cdot|\cdot)$ denotes their \emph{conditional
entropies} interpreted as \emph{surprise} and \emph{uncertainty}, respectively.

\noindent\textbf{Results:} It is first inferred that to minimize the metric an
individual can learn about $R$ (and store that knowledge in $M$) or can
restrict $R$ to the already known $M$. Then, it is concluded that
\emph{insistence on sameness} is a manifestation of the latter. Moreover, it is shown that the proposed framework:
\begin{itemize}
\item Helps to quantify the concepts of \emph{surprise, uncertainty, sensory
overload and deprivation, anxiety, comfort zone, disappointment,
disorientation, pedantry, rigidness, observance} or \emph{aberrant precision}.

\item Leads to a list of guidelines for learning therapies and daily care
routines, and allows them to be defined as optimization algorithms and
implemented as programs for \emph{robotic live-in caregivers}.

\item Can be validated with the help of a \emph{Turing test}-like approach
that requires no experiments involving individuals with autism.
\end{itemize}

\noindent\textbf{Conclusion:} The framework -- if positively validated -- will
provide advantages of both theoretical and practical importance: it explains
the insistent on sameness as a consequence of cognitive restrictions and
offers formal foundations and design guidelines for therapies aimed at
improving\emph{ self-reliance} of individuals with autism in\emph{ basic activities of daily living}.

\end{abstract}


\section{Introduction}

Etiology and pathogenesis of autism remain unclear as does its definition \citep{Wigg:2009,Amer:2013,Davi:2015,Saue:2021,Volk:2021,Jour:2023,Shir:2024,Assa:2024,Mahe:2025}. There is no a consensus about the behavior traits that are associated with individuals with autism either \citep{Mott:2020,Chap:2021,Pero:2023}. While behavior, in general, appears to be driven by a variety of templates and patterns (see \emph{e.g.} \cite{Lai:2023} and \emph{cf.} \cite{Bail:2024}), we focus on \emph{insistence on sameness}, a phenomenon referred also to as \emph{stereotyped, repetitive}, and the \emph{autistic-like behavior} or the \emph{autism-like trait} \citep[\emph{cf. e.g.} ][]{Ulja:2017,Volk:2021,Lyu:2024,Kamp:2024} observed in non-verbal low-functioning individuals \citep[the term \emph{low-functioning} is used to denote \emph{severe} (\emph{profound}) instances of the \emph{Level 3 of Autism Spectrum Disorder}, \emph{cf.}][]{Amer:2013,Bett:2015,Wang:2022,Sing:2023}.

We propose a new insight into that phenomenon and examine its resemblance to a problem of \emph{constrained minimization of uncertainty}, where the following \emph{entropy metric} \citep[see \emph{e.g.}][]{Mack:2003}:
\begin{equation}
D_{H}\left(  R,M\right)  =H\left(  R|M\right)  +H\left(  M|R\right)  ,
\label{eq:entropy_metric}%
\end{equation}
is used as a \emph{measure of uncertainty}. Random variables, $R$ and $M$, represent the real-world environment and the memory of an individual, respectively. The terms $H\left(  R|M\right)  $ and $H\left(  M|R\right)$ are \emph{conditional entropies} that quantify the levels of \emph{surprise} and \emph{uncertainty}.

The motivation for such an approach is to devise a mathematically rigorous framework that offers an \emph{explanation} of the insistence on sameness which is more precise and less arbitrary than the now dominant language-based descriptions \citep[\emph{cf. e.g.}][]{Amer:2013,Mott:2020,Chap:2021,Volk:2021,Wang:2022}, enables formulation of learning therapies as solutions to a constrained optimization problem and facilitates their implementations as programs for \emph{e.g.} robot-based daily life assistants.


%
\defcitealias{Amer:2013}{Amer.~Psych.~Assoc}
\defcitealias{Davi:2015}{Davis\,\&\,Plaisted-Grant}
\begin{table}[th]
\centering
\label{tab:matrix}
\begin{tabular}{p{1.4in}p{4.6in}}
Work & Related scope \\\hline\hline
\citeauthor{Mack:2003}  &  An accessible information theory textbook, where basic definitions of the \emph{entropy metric, conditional and relative entropies} are introduced and derived.\\
\citeauthor{Fris:2010}  &  A review paper that discusses a hypothesis that the \emph{free-energy principle} can be a basis for a universal brain theory.\\
\citetalias{Amer:2013}  &  A diagnostic manual that presents diagnostic criteria, verbal descriptions of behavior traits associated with autism (incl. \emph{insistence on sameness}); definitions of the \emph{autism spectrum disorder severity levels 1-3}.\\
\citeauthor{Laws:2014}  &  Introduction, analysis and early experimental verification of a link between \emph{free-energy principle} and \emph{aberrant precision} phenomenon in autism.\\
\citeauthor{Gali:2016}  &  An attempt to create a computational model of autism perception based on mostly verbal description of the functional \emph{theory of mind}. \\
\citetalias{Davi:2015}  &  Discussion of a possibly multifold (positive and negative) impact of noise on autism traits and the associated cognitive abilities.\\
\citeauthor{Palm:2017}  &  Reviews Bayesian approaches to autism. Introduces a model based on assumption that underlies a significant impact of the sensory information on building and updating the internal representation of environment.\\
\citeauthor{Ulja:2017}  &  A study that links \emph{insistence on sameness} with \emph{self-regulation} in autism. It also verifies experimentally its relationship with \emph{effortful control} and \emph{anxiety} phenomena.\\
\citeauthor{Mott:2020}  &  A proposal of several remedies to the problem of multifold increase in autism diagnoses, incl. narrowing its definition and combining the experience of clinicians with qualitative features. \\
\citeauthor{Volk:2021}  &
A comprehensive, extensive and detailed compendium describing autism spectrum disorder-related topics; in particular, \emph{low-functioning individuals} and \emph{insistence on sameness}.\\
\citeauthor{Wang:2022}  &  A survey presenting \emph{state-of-the-art} results in the areas of autism diagnosis and intervention. Prioritizes \emph{functional-based} therapeutic tools.\\
\citeauthor{Rams:2023}  &  A review paper which discusses in a formalized fashion the universality of \emph{free-energy principle} and explains its relationship to other principles and laws of physics and information theory. \\\hline
\end{tabular}
\caption{Works that led to the development of the framework.}%
\end{table}

\noindent The framework is inspired by a variety of results and observations from information theory, neuroscience and psychiatry (see Table \ref{tab:matrix}) and its contribution is twofold, \emph{theoretical} (a relatively simple and verifiable explanation of \emph{insistence on sameness} as a consequence of cognitive restrictions of a person with autism) and \emph{practical} (a set of design guidelines for therapies focused on \emph{low-functioning non-verbal} individuals), and may lead to a more precise definition of the disorder and better personalized therapies.

Clearly, there are some limitations of the framework: it is a \emph{functional} model, \emph{i.e.}, it does not define the underlying biological implementation. It also requires new and carefully designed personalized validation experiments to be developed. In turn, limiting the scope of the framework to low-functioning non-verbal individuals with autism is a deliberate attempt to make it both formal and accurate.

In the following sections we present the framework, examine its basic properties and discuss its relations with insistence on sameness. Then, we devise the guidelines for learning therapies aimed at increasing autonomy of persons with autism, formulate a therapy as an optimization problem and discuss two framework validation approaches. Finally, we conjecture that \emph{insistence on sameness} is a special case of a general minimizing uncertainty pattern \citep[\emph{cf.}][]{Fris:2010} and conclude the work with the proposal of a new definition of autism.

\section{Methods}
The framework consists of three components: the entropy metric (\ref{eq:entropy_metric}), a stimuli processing loop in Fig. \ref{fig:processing-loop}, and a set of Assumptions \ref{ass:first}-\ref{ass:last} that represent the cognitive constraints of a person with autism.

\subsection{Entropy metric}

\label{sec:entropy_metric}

The metric (\ref{eq:entropy_metric}) measures an \emph{entropy distance} between two discrete random variables $R$ and $M$. To illustrate its basic properties, we will consider two opposite scenarios:

\begin{enumerate}
\item $R$ fully determines $M$ (and \emph{vice versa}). In this case knowledge of one variable removes the entire uncertainty about the other and thus both conditional entropy terms are zero (and so is the metric),
\[
D_{H}\left(  R,M\right)  =\underset{=0}{\underbrace{H\left(  R|M\right)  }%
}+\underset{=0}{\underbrace{H\left(  M|R\right)  }}=0.
\]

\item Both $R$ and $M$ are independent of each other. Knowing either of them carries no information about the other (so they maintain their original uncertainties):
\[
D_{H}\left(  R,M\right)  =\underset{=H\left(  R\right)  }{\underbrace{H\left(R|M\right)  }}+\underset{=H\left(  M\right)  }{\underbrace{H\left(M|R\right)  }}=H\left(  R\right)  +H\left(  M\right).
\]

\end{enumerate}

Hence, in order to reduce the entropic distance between $R$ and $M$ one can learn about $R$ (and store that knowledge in $M$), or constrain $R$ to the already known $M$.

\subsection{Stimuli processing loop}

The loop (see Fig. \ref{fig:processing-loop}) serves as a model of stimuli processing. Its two phases, \emph{perception}, and \emph{prediction} are associated with the components of the metric, $H\left(  R|M=m_{n-1}\right) $ and $H\left(  M|R=r_{n}\right)  $, that quantify how \emph{surprise} and \emph{uncertainty} are affected by the incoming stimulus $r_{n}$. If levels of surprise or uncertainty exceed the respective thresholds, $T_{so}$ or $T_{a}$, a \emph{sensory overload} or \emph{anxiety} state can occur.

\usetikzlibrary {calc, positioning, shapes.misc, shapes.geometric, arrows.meta, graphs, backgrounds, shadows}

\tikzset{shadow/.style = {scale=1, shadow xshift=.5ex, shadow yshift=-.5ex,
  opacity=.5, fill=black!50}}

\tikzset
{
        starter/.style =
        {
                drop shadow,
                fill = white,
                circle,
                minimum size = 5mm,
                very thick,
                draw = green!20!blue!75!black!50,
                text = green!20!blue!75!black!75,
                text width = 1cm, align = center,
                font = \bf,
        }
}
\tikzset{terminal/.style = {
      drop shadow,
      fill = white,
      rectangle, rounded corners = 1mm,
      minimum size = 10mm,
      very thick,
      draw = green!20!blue!75!black!50,
      text = green!20!blue!75!black!75,
      text width = 2cm, align = center,
      font = \bf,
}}
\tikzset{nonterminal/.style = {
        drop shadow,
        fill = white,
        rectangle, minimum size = 10mm, rounded corners = 1mm,
        text width = 20mm, align = center,
        very thick, draw = black!50,
        text = black,
        font = \bf,
}}
\tikzset{setterminal/.style = {
        fill = white,
        rectangle, minimum size = 14mm, rounded corners = 1mm,
        text width = 24mm, align = center,
        densely dashed,
        semithick, draw = black!25,
        text = black,
        font = \bf,
}}
\tikzset{dummy/.style = {
        rectangle
}}

\tikzset{decision/.style = {
        drop shadow,
        diamond, rounded corners,
        text centered, text width = 21mm,
        text = black,
        font = \bf,
        fill = white,
        align = center,
        very thick, draw = black!50,
}}
\begin{figure}[tbh]
\centering
\begin{tikzpicture}[scale=1,node distance = 5mm and 5mm, > = {Stealth[round]},thick]
  \node (perc) [decision]      {Perception \vspace*{1mm} \hrule \vspace*{1mm} \color{green!20!blue!75!black!75}{Surprise}};
  \node (m)    [nonterminal, above = of perc] {$m_{n-1}$};
  \node (r)    [starter, left = of perc]  {$r_{n}$};
  \node (sov)  [terminal, below = of perc]    {Sensory overload};
  \node (m_n)  [nonterminal, right = of perc] {$m_{n} = \rho(r_n)$};
  \node (pred) [decision, right = of m_n]   {Prediction \vspace*{1mm} \hrule \vspace*{1mm} \color{green!20!blue!75!black!75}{Uncertainty}};
  \node (anx)  [terminal, below = of pred]    {Anxiety};

  \draw [->, rounded corners=1mm] (pred.north) |- (m.east)
        node (fb) [below left, midway, text = black] {$n = n + 1$};
  \draw [->]  (m.south) -- (perc.north);
  \draw [->]  (r.east) -- (perc.west);
  \draw [->]  (perc.east) -- (m_n.west);
  \draw [->]  (m_n.east) -- (pred.west);
  \draw [->, draw = green!20!blue!75!black!75] (perc.south) -- (sov.north)
        node [left, midway] {$ H(R|M = m_{n-1}) > T_{so}$};
  \draw [->, draw = green!20!blue!75!black!75] (pred.south) -- (anx.north)
        node [left, midway] {$ H(M|R = r_n) > T_{a}$};
\end{tikzpicture}

 \caption{Stimuli processing loop.}%
\label{fig:processing-loop}%
\end{figure}
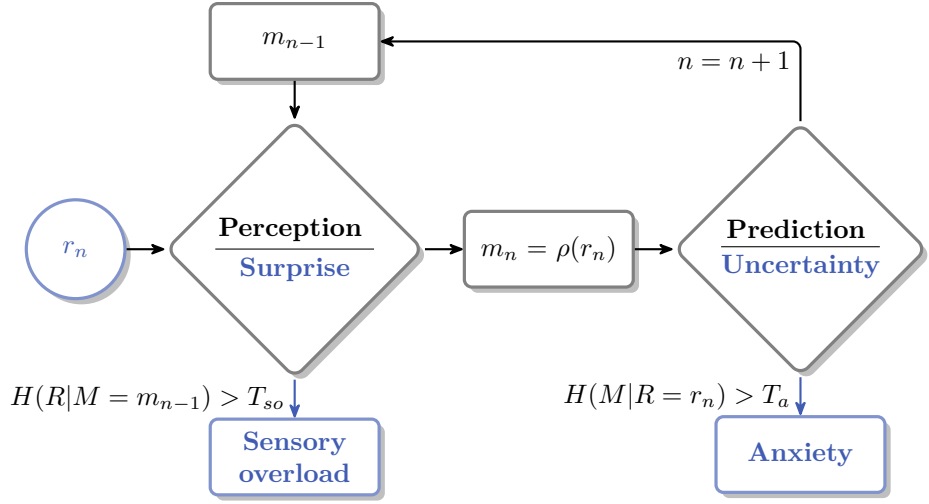

\subsection{Assumptions}

To specify the cognitive constraints of a non-verbal low-functioning individual with autism, we assume that:

\begin{enumerate}
[label=\textbf{A\arabic{enumi}:}, ref=A\arabic{enumi}]

\item \label{ass:first}The \emph{stimuli}, $r_{n},n=0,1,\ldots$, represent a perceived environment and form sequences. The mixture of all sequences is denoted by $R$.

\item \label{ass:associative_memory}A memory, $M$, is capable of storing sequences and retrieving their elements, $m_{n}, n=0,1,\ldots$.

\item \label{ass:nn_association}Each memory item, $m_{n}=\rho(r_{n})$, is a result of a classification of a single stimulus, $r_{n}$, by a \emph{nearest neighbor algorithm}.

\item \label{ass:last}Both \emph{sensory overload} and \emph{anxiety} thresholds, $T_{so}, T_{a} > 0$, are random variables.
\end{enumerate}

\subsection{Comments on assumptions}

\label{sec:assumption_remarks} 
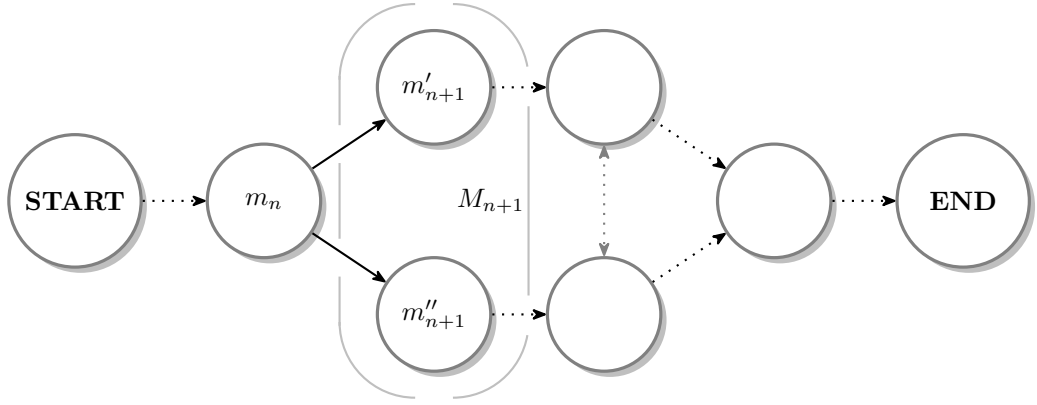
\begin{figure}[tbh]
\centering
\begin{tikzpicture}[scale=1, transform shape,
      node distance= 1.5cm and 2.25cm,
      > = {Stealth[round]},thick,on grid,initial/.style    ={}]
\usetikzlibrary {calc, positioning, shapes.misc, shapes.geometric, arrows.meta, graphs, backgrounds, shadows}

\tikzset{shadow/.style = {scale=1, shadow xshift=.5ex, shadow yshift=-.5ex,
font = \bf,  opacity=.5, fill=black!50}}

\tikzset{every node/.style={circle, fill = white}}
\tikzset{state/.style={minimum size=1.5cm,draw=black, circle, drop shadow,         very thick, draw = black!50,
text = black,
font = \bf,
}}
\tikzset{strts/.style={minimum size=1.75cm,draw=black, circle, drop shadow,     very thick, draw = black!50,
text = black,
font = \bf,
}}
\tikzstyle{dotted}= [->, dash pattern=on \pgflinewidth off 1mm, draw=black, thick]
\tikzstyle{ddotted}= [<->, dash pattern=on \pgflinewidth off 1mm, draw=gray, thick]
\tikzset{gryline/.style={->,relative=false, draw=black}}
\tikzset{redline/.style={->,relative=false, draw=blue}}
\tikzset{brdr/.style={relative=false, draw=lightgray}}

\node[strts]   (S)                          {START};

\node[state]   (MN)   [right = of S, xshift=.25cm]        {$m_{n}$};
\node[state]   (M1N)  [above right = of MN] {$m'_{n+1}$};
\node[state]   (M2N)  [below right = of MN] {$m''_{n+1}$};
\node[state]   (M1)   [right = of M1N]      {};
\node[state]   (M2)   [right = of M2N]      {};
\node[state]   (M3)   [below right = of M1] {};

\node[strts]   (E)    [right = of M3, xshift=.25cm]       {END};
\node          (MM)   [below = of M1N, xshift=.75cm]      {$M_{n+1}$};

\path (S)    edge [dotted]  (MN)
      (MN)   edge [gryline] (M1N)
      (MN)   edge [gryline] (M2N)

      (M1N)  edge [dotted]  (M1)
      (M2N)  edge [dotted]  (M2)
      (M1)   edge [ddotted] (M2)

      (M1)   edge [dotted]  (M3)
      (M2)   edge [dotted]  (M3)
      (M3)   edge [dotted]  (E)
;
\draw[brdr] (5,2.6) arc (90:10:1cm);
\draw[brdr] (4.5,2.6) arc (90:180:1cm);

\draw[brdr] (6, 1.25) -- (6, -1.25);

\draw[brdr] (3.5, 1) -- (3.5, 1.65);
\draw[brdr] (3.5, 0.5) -- (3.5, -0.5);
\draw[brdr] (3.5, -1.0) -- (3.5, -1.65);

\draw[brdr] (5,-2.6) arc (-90:-10:1cm);
\draw[brdr] (4.5,-2.6) arc (-90:-180:1cm);

\end{tikzpicture} \caption{An instance of a graph composed of memorized stimuli sequences, $m_{n} = \rho(r_{n}),n=0,1,\ldots$ with a set $M_{n+1} = \{m_{n+1}^{\prime}, m_{n+1}^{\prime\prime}\}$ consisting of a pair of alternative events.}
\label{fig:sequences}
\end{figure}

Each stimulus, $r_{n}$, can be interpreted as a "\emph{snapshot}" of the environment and $m_{n}=\rho\left(  r_{n}\right)  $ as its counterpart perceived by the senses and located in memory $M$ \citep[\emph{cf. e.g.}][]{Nobr:2022}. The sequences in the memory are represented by chains/directed graphs (see Fig. \ref{fig:sequences}).

\begin{remark}
\label{rem:unified_stimuli}All stimuli are treated in a unified way, as raw (random)\ signals, and their sources (\emph{e.g.}\ external/exogenous\ or internal/endogenous) are not distinguished. Moreover, their semantics (\emph{i.e.}\ the non-tangible, abstract, cultural meanings and associations) are not available. Thus, only their perceptible properties are taken into account and treated as equally important; \emph{i.e.} all their abstract properties and relations \citep[see \emph{e.g.}][]{Back:2014}, incl. \emph{importance}, \emph{hierarchy}, \emph{being a part/whole} are ignored.
\end{remark}

\begin{example}
An abstract notion of \emph{nothing} cannot be represented by any perceptible stimulus. This observation can also be derived from the namesake property of a \emph{nearest neighbor algorithm} (\emph{cf.} Assumption
\ref{ass:nn_association}), where the classification routine always selects the closest remembered (known) item, regardless of how "far" it is in terms of a distance function.
\end{example}

\begin{remark}
\label{rem:zero-order-inference}Assumptions \ref{ass:associative_memory}-\ref{ass:nn_association} define a cognitive model with the inference abilities limited to replication of the memorized stimuli and their sequences. They also imply a non-vanishing (persistent) character of the memory.
\end{remark}

\begin{remark}
\label{rem:random_thresholds}Defining the thresholds, $T_{so},T_{a}$, as random variables allows them to be unknown, varying in time, specific for an individual and driven by the factors not included in the model \citep[\emph{cf. e.g.}][]{Kim:2023,Lai:2023}.
\end{remark}

\subsection{Analysis\label{sec:analysis}}

The goal of this section is to inspect how the metric changes during \emph{perception} and \emph{prediction} phases; \emph{cf.} the processing loop in Fig. \ref{fig:processing-loop}. The amount of \emph{surprise} in the former, given the memorized stimulus, $m_{n-1}$, is
\begin{equation}
D_{H}\left(  R,M|M=m_{n-1}\right)  =H\left(  R|M=m_{n-1}\right)+\underset{=0}{\underbrace{H\left(  M=m_{n-1}|R\right)  }},
\label{eq:perception_entropy}%
\end{equation}
while the level of \emph{uncertainty} in the latter, given the incoming stimulus, $r_{n}$, is
\begin{equation}
D_{H}\left(  R,M|M=r_{n}\right)  =\underset{=0}{\underbrace{H\left(R=r_{n}|M\right)  }}+H\left(  M|R=r_{n}\right)  ,
\label{eq:prediction_entropy}
\end{equation}
so that during \emph{perception} we need to consider only the first term, $H\left(  R|M=m_{n-1}\right)  $, and only the other, $H\left(  M|R=r_{n}\right)  $, during \emph{prediction}. We examine them in the following three cases; \emph{cf.} Sec. \ref{sec:entropy_metric}:

\begin{enumerate}
[label=\textbf{C\arabic{enumi}:}, ref=C\arabic{enumi}]

\item The first, \emph{deterministic}, takes place when in \emph{perception} phase knowledge of $m_{n-1}$ determines $r_{n}$ so that there is no \emph{surprise}, and when there is no \emph{uncertainty} during
\emph{prediction}, because knowledge of $r_{n}$ determines $m_{n+1}$. Then, both terms are zero:
\begin{equation}
H\left(  R=r_{n}|M=m_{n-1}\right)  =0\text{ and }H\left(  M=m_{n+1}|R=r_{n}\right)  =0.
\label{eq:deterministic}
\end{equation}

\item \label{case:set} The second case occurs when, in the \emph{perception} phase, $m_{n-1}$ determines a set $R_{n}$ of possible stimuli or, when in \emph{prediction}, the set of memorized possibilities $M_{n+1}$ occurs given the stimulus $r_{n}$ (\emph{cf.} Fig. \ref{fig:sequences}). The respective metric terms are:
\begin{equation}
H\left(  R=R_{n}|M=m_{n-1}\right)  =H\left(  R_{n}\right)  \text{ and }H\left(  M=M_{n+1}|R=r_{n}\right)  =H\left(  M_{n+1}\right)  .
\label{eq:sets}
\end{equation}

\item \label{case:unknown}The last case encompasses, amongst others, the learning-like situations when $r_{n}$ is a new (unknown)\ stimulus, so that the current knowledge of $m_{n-1}$ carries no information about it during \emph{perception} and the stimulus itself does not reduce the level of \emph{prediction uncertainty}:
\begin{equation}
H\left(  R|M=m_{n-1}\right)  =H\left(  R\right)  \text{ and }H\left(M|R=r_{n}\right)  =H\left(  M\right)  .
\label{eq:unknown}
\end{equation}

\end{enumerate}

\noindent It should be noted here that because the sets of known options, $R_{n}$ and $M_{n+1}$, are subsets of $R$ and $M$, the amounts of \emph{surprise} or \emph{uncertainty} in \ref{case:set} seem to always be
smaller than in \ref{case:unknown} (when unknown stimuli occur). Nevertheless, for some realizations of $m_{n-1}$ and $r_{n}$ the following inequalities may hold \citep[see \emph{e.g.}][Ch. 8]{Mack:2003}:
\begin{equation}
H\left(  R_{n}|M=m_{n-1}\right)  \geq H\left(  R\right)  \text{ and\ }H\left(M_{n+1}|R=r_{n}\right)  \geq H\left(  M\right),
\label{eq:sets_vs_unknown}
\end{equation}
and hence the \emph{surprise} and \emph{uncertainty} levels in (\ref{eq:sets}) can be larger than in (\ref{eq:unknown}).

\begin{example}
\label{ex:disorientation} Such a situation can take place when, in \emph{prediction} phase, the set $M_{n+1}$ consists of multiple options, $\{m_{n+1}^{\prime},\allowbreak m_{n+1}^{\prime\prime},\allowbreak
\ldots,\allowbreak m_{n+1}^{(\nu)}\}$, some $\nu$, that are equally probable; \emph{cf.} Fig. \ref{fig:sequences}. The amount of \emph{uncertainty} is then increased (predictions are the least efficient) and can easier lead to \emph{disorientation} and \emph{anxiety} (especially when $\nu$ is large).
\end{example}

\noindent So far, we have conditioned the metric terms on the last realizations of the stimulus $r_{n}$ (or on the latest memory item $m_{n-1}$), as if the amounts of \emph{surprise }and \emph{uncertainty }are independent of the past (like in Markov chains \citep[see \emph{e.g.} ][]{Ekro:1993}). However, to better comply with Assumptions \ref{ass:associative_memory}-\ref{ass:nn_association} and Remarks \ref{rem:unified_stimuli}-\ref{rem:zero-order-inference}, one should take into account the previous events as well.

\begin{example}
\label{ex:disappointment} If, for instance, in the \emph{perception} phase, we condition the \emph{surprise} on the set $M_{n-1}=\left\{  m_{n-1},m_{n-2},\ldots\right\}  $ of all previous items in a sequence, then the term $H\left(  R|M=M_{n-1}\right)  $ can quantify a \emph{disappointment} that the sequence does not continue in a way it has been memorized.
\end{example}

\subsubsection{Classification}

The influence of the \emph{classification} phase (\emph{i.e.} the assignment $m_{n}=\rho(r_{n})$ in Fig.~\ref{fig:processing-loop}) has so far been ignored. This is because it resembles the impact of \emph{prediction}. First, observe that the metric reduces to the same term; \emph{cf.} (\ref{eq:prediction_entropy}):
\[
D_{H}\left(  R=r_{n}, M\right)  =\underset{=0}{\underbrace{H\left(R=r_{n}|M\right)  }}+H\left(  M|R=r_{n}\right)  .
\]
Next, note that if the stimulus, $r_{n}$, is already known, then it fully determines the corresponding $m_{n}$ (\emph{cf.} Assumption \ref{ass:associative_memory} and Remarks \ref{rem:unified_stimuli}%
-\ref{rem:zero-order-inference}). In turn, if $r_{n}$ is new, it carries no information about the memory. Hence, \emph{cf.} (\ref{eq:deterministic}) and (\ref{eq:unknown}):
\[
H\left(  M|R=r_{n}\right)  =\left\{
\begin{array}
[l]{l}%
0\\
H\left(  M\right)
\end{array}
\text{ if }r_{n}\text{ is }%
\begin{array}
[l]{l}%
\text{known,}\\
\text{new.}%
\end{array}
\right.
\]

\begin{remark}
\label{rem:nn} Although the presence of randomness, even in a fixed and arranged environment, seems to be inevitable \citep[\emph{cf. e.g.}][]{Davi:2015,Walk:2023}, the modified environments and/or the sequences can still be perceived as known as long as those random changes do not affect the results of classification; \emph{cf.} Remark \ref{rem:unified_stimuli}.
\end{remark}

\section{Results}

\subsection{Analogies with insistence on sameness\label{sec:behavior}}

In case of non-verbal low-functioning individuals with autism, both \emph{sensory overload} and \emph{anxiety} can lead to violent, (self-)aggressive behaviors (sometimes labeled as \emph{meltdowns} or
\emph{tantrums} \citep{Volk:2021}). Insistence on sameness can be seen as a set of actions aimed at decreasing the risk of these events, either by observing (learning) the environment (\emph{i.e.} making $M$ better
approximate $R$), or by constraining it to the already known one (that is, making $R$ better resemble $M$). For example \citep[\emph{cf.} ][]{Ulja:2017}:

\begin{enumerate}
[label=\textbf{B\arabic{enumi}:}, ref=B\arabic{enumi}]

\item \label{beh:pedantry} \emph{Pedantry}, together with a \emph{routinized} and \emph{repetitive} behavior (incl. strict following only known sequences of activities and keeping the known environments unchanged; \emph{cf.} an \emph{aberrant precision} in \citep{Laws:2014}), are the means to reduce the levels of surprise or uncertainty; see Ex. \ref{ex:disorientation}, \ref{ex:disappointment} and Remark \ref{rem:nn}.

\item \emph{Ritualistic} and \emph{stereotyped} behaviors stem from the lack of the ability to ignore items; see Remarks \ref{rem:unified_stimuli}-\ref{rem:zero-order-inference} and \emph{cf.} Guideline \ref{lim:correlative}.

\item \emph{Restrictive} and\emph{\ rigid} behavior includes actions like forcing deterministic scenarios in order to further control \emph{surprise} and \emph{uncertainty} (and to effectively zero their levels as in (\ref{eq:deterministic})), to avoid \emph{disappointments} and to select a preferred (the most frequent) option in multiple option cases; \emph{cf.} (\ref{eq:sets}), (\ref{eq:sets_vs_unknown}) and Ex. \ref{ex:disappointment}.

\item \emph{Vigilance} and \emph{observance} reduce the number of options, that is, the uncertainty level caused by them in (\ref{eq:sets}); see Ex. \ref{ex:disorientation}.

\item \label{beh:aversion} Aversion to learning is a way to avoid potentially high surprise/uncertainty levels caused by new options or new stimuli; \emph{cf.} (\ref{eq:sets})-(\ref{eq:sets_vs_unknown}).
\end{enumerate}

\subsection{Therapeutic guidelines}

\label{sec:guidelines} In practical terms, the goal of the learning therapy is twofold:

\begin{enumerate}
[label=\textbf{L\arabic{enumi}:}, ref=L\arabic{enumi}]

\item \label{goal:auto} Increasing autonomy of an individual (by learning new vital activities in a supervised way),

\item \label{goal:explore} Reduction of aversion to exploration (by controlling uncertainty of the new environments designed for semi-supervised and unsupervised activities).
\end{enumerate}

Assumptions \ref{ass:first}-\ref{ass:last}, Remarks \ref{rem:unified_stimuli}-\ref{rem:random_thresholds}, and the subsequent analysis in Sec. \ref{sec:analysis}, imply that the therapy is a long-term incessant process with several limitations on how it can be implemented (in particular, they exclude rule-based learning; \emph{cf.} Remarks \ref{rem:unified_stimuli}-\ref{rem:zero-order-inference} and \citep{Bett:2015,Gali:2016}):

\begin{enumerate}
[label=\textbf{G\arabic{enumi}:}, ref=G\arabic{enumi}]

\item \label{lim:tangible}Only tangible objects and their realistically simulated counterparts should be used in learning environments and in sequences of activities.

\item \label{lim:abstracts} Artifacts such as pictures (especially, in simplified forms of drawings or labels), but also spoken words and sentences, should \emph{a priori} be assumed as unrelated with the tangible items they represent or refer to. Hence, if necessary, these relations need to be learned separately. Abstract notions, like being a child or a parent, a teacher or a caretaker, and relations like love, trust or friendship, are not available either.

\item \label{lim:correlative}The meaning of tangible items depends on the context (created by other items and/or their past sequences; \emph{cf.} Remark \ref{rem:zero-order-inference}). Such a correlative-like reasoning leads to "\emph{cum/post hoc ergo propter hoc}" fallacies; see Example \ref{ex:delayed_gratification}.

\item Remembering the exact sequences (of events and activities; \emph{cf.} Remarks \ref{rem:unified_stimuli} and \ref{rem:zero-order-inference}) implies they are treated as a whole and that "\emph{shortcuts}" in sequences or other "\emph{by-the-way}" activities will likely be perceived as new (and will increase levels of surprise and/or uncertainty); \emph{cf.} Ex. \ref{ex:disappointment} and \citep{Laws:2014,Noel:2025}.

\item \label{lim:forks}Multiple choice situations (branches in sequences; see Fig. \ref{fig:sequences}) introduce uncertainty even if the environment remains unchanged; \emph{cf.} Cases \ref{case:set}-\ref{case:unknown}, and Exs. \ref{ex:disorientation}-\ref{ex:disappointment} and \ref{ex:talisman}.
\end{enumerate}

\begin{example}
\label{ex:talisman} In order to reduce the risk of anxiety induced by predictions in multiple option/choices situations, several countermeasures could be applied: making all sequences distinct (with the help of
\emph{amulet- or talisman-like} artifacts that will allow to distinguish the otherwise same activities), or locating the branches at the sequence beginnings (to reduce these uncertainties as early as possible), or limiting the numbers of possible options in them.
\end{example}

\begin{example}
\label{ex:delayed_gratification} An instance of an immediate reward will likely turn into a part of a routine while its delayed version will become a part of (possibly unrelated) sequence it will occur in. This is because the delayed gratification concept relies on an abstract concept of an award and its causal relation with the past events.
\end{example}

\subsection{Self-stimulation activities and comfort zone}

\noindent The guidelines \ref{lim:tangible}-\ref{lim:forks} are aimed at reducing risk of tantrums during learning/therapy. However, in case of non-verbal individuals, it is difficult to assess whether the levels of
surprise and/or uncertainty are sufficiently low (w.r.t. their thresholds $T_{so}\text{ and }T_{a}$) to safely start a new sequence of activities. The problem is of special importance at the beginning of the therapy when the environments and activities the individuals are familiar with (incl. their routines stored in $M$) are not known. Nevertheless, if the stimuli are treated in a unified way, so that their exogenous or endogenous nature is not distinguished (\emph{cf.} Assumption \ref{ass:first} and Remark \ref{rem:unified_stimuli}), we can conjecture that \citep[\emph{cf.}][]{Davi:2015}:

\begin{conjecture}
\emph{Self-stimulation} activities (incl. \emph{"flapping, waving, finger wiggling, mouth opening, orofacial movements, head nodding"} \citep[see][]{Volk:2021}), are a sign of low levels of perceived surprise and uncertainty.
\end{conjecture}

\noindent The rationale behind this claim is following: without external stimuli (in a state of \emph{sensory deprivation}), the internal ones become dominant. If sufficiently random, they can carry an amount of entropy large enough to exceed either sensory overload or anxiety threshold. \emph{Self-stimulations} can therefore (in case of low-functioning individuals, in particular) be a way to generate known (zero- or low-entropy) stimuli that keep the surprise/uncertainty levels below these thresholds (\emph{i.e.}, inside a 'comfort zone').

\begin{remark}
  \label{rem:video-compression}
  One of the Reviewers raised a question about the role of \emph{white noise}. That is an interesting and open problem as one can consider two potentially opposite, \emph{negative} and \emph{positive}, effects:
  \begin{enumerate}[label=\textbf{N\arabic{enumi}:}, ref=N\arabic{enumi}]
    \item Presence of a white noise in stimuli increases their conditional entropy and the risk of \emph{sensory overload}. It can affect the results of the classification and increase the risk of \emph{anxiety} as well; \emph{cf.} Remark \ref{rem:nn}.
    \item In turn, as pointed out by \cite{Davi:2015}, noise can help avoid getting stuck in local minima during optimization and, for instance, enable generalization (abstraction) abilities. In our framework, such effect can be associated with jumping between the known sequences or finding the shortcuts within them.
\end{enumerate}
\noindent  White noise, if present in exogenous visual or aural stimuli, could be detected with the help of basic audio-video codecs (the more noisy environment, the less effective entropy coders; see \citet[Ch. 1]{Mack:2003}).
\end{remark}
\subsection{Learning therapy as optimization problem\label{sec:learning}}

In the framework's terms, learning a low-functioning person with autism is equivalent to presenting new stimuli and their sequences. Therefore, a learning therapy with the goals \ref{goal:auto}-\ref{goal:explore}, can be formulated as the following optimization problem:

\begin{problem}
\label{pr:learning}Given an initial state of a memory $M$, find $R$ such that
\begin{equation}
\underset{R}{\arg\max}\ I\left(  R;M\right)  ,
\label{eq:objective_mutual_information}%
\end{equation}
subject to%
\[
H\left(  R|M=m_{n-1}\right)  <T_{so}\text{ and }H\left(  M|R=r_{n}\right)<T_{a},
\]
given Assumptions \ref{ass:first}-\ref{ass:last} and the processing loop in Fig. \ref{fig:processing-loop}.
\end{problem}

The \emph{objective function }(\ref{eq:objective_mutual_information}) is the \emph{mutual information} of $R$ and $M$, and is related to the entropy metric in (\ref{eq:entropy_metric}) via identity $D_{H}\left(  R, M\right)  =H\left(R,M\right)  -I\left(  R;M\right)  $, where $H\left(  R,M\right)  $ is the joint entropy of $R$ and $M$; see \citet[Ch. 8]{Mack:2003} and \citet{Fris:2010,Rams:2023}. The practical value of such a formulation is that it allows learning therapies and daily care routines to be defined as optimization algorithms and implemented as programs
for \emph{robotic live-in caregivers}.

\section{Discussion}

\subsection{Validation\label{sec:Model_validation}}

One can consider the following two (complementary) approaches to validate/reject the framework's assumptions and the resulting analogies:

\begin{enumerate}
[label=\textbf{V\arabic{enumi}:}, ref=V\arabic{enumi}]

\item \label{ver:traditional}A traditional (direct) variant, where the realistic environments and interactively generated sequences of new activities are presented through a "digital window" (\emph{e.g.}, a wall-like display with a haptic interface). The intensity (and the pace the sequences are presented with) depends, in accordance to recommendations \ref{lim:tangible}-\ref{lim:forks}, on the levels of surprise and anxiety assessed from \emph{e.g.} the observed behavior or from a frequency of interactions between the individual and the generated world. The long-term therapeutic effects, like increased autonomy in these environments (with possible help of appropriately programmed robotic live-in caregivers), is evaluated in a traditional way (\emph{e.g.} in a form of progress questionnaires) by parents and therapists.
\end{enumerate}

\noindent This is a standard approach that requires an access to individuals with autism living in their homes (or other familiar environments; \emph{cf.} the analogies \ref{beh:pedantry}-\ref{beh:aversion}). Getting permissions for such arrangements is usually difficult (especially at early validation stages). Moreover, in order to satisfy \emph{reproducibility} conditions, each experiment needs a tedious and time-consuming individual setup (\emph{cf.} Assumption \ref{ass:associative_memory} and Ex. \ref{ex:disappointment}). Therefore, as an alternative (as a preliminary test phase) we propose:

\begin{enumerate}
[label=\textbf{V\arabic{enumi}:}, ref=V\arabic{enumi}]
\setItemnumber{2}%

\item A \emph{Turing test}-like approach, in which the framework is used to create avatars ("digital twins") residing in realistic virtual environments, where their simulated behavior is observed and validated by therapists or by early-detection computer tools; \emph{cf.} \citep{Dill:2023} and \citep{Pero:2023}, respectively.
\end{enumerate}

\begin{remark}
\label{rem:opportunity_cost} Validation methods which are useful to assess higher-functioning individuals and which (implicitly) assume an individual is able to perceive abstracts and operate on them, \emph{e.g.} by examining their propensity to ignore \emph{opportunity costs} \citep[see \emph{e.g.}][]{DaSi:2025}, cannot be used here.
\end{remark}

\subsection{Final remarks}

\noindent We conclude the work with a claim that reducing uncertainty is not specific to autism.

\begin{conjecture}
The phenomenon of \emph{insistence on sameness} in autism is a special case of the generic behavior pattern described in \citep{Fris:2010} where "\emph{Adaptive agents must occupy a limited repertoire of states and
therefore minimize the long-term average of surprise associated with sensory exchanges with the world} [and that] \emph{action under the free-energy principle reduces to suppressing sensory prediction errors that depend on predicted (expected or desired) movement trajectories.}"
\end{conjecture}

\noindent Therefore, a distinctive character of \emph{insistence on sameness} should rather be attributed to cognitive limitations of a person with autism. Thus, if correct, the claim corroborates the following definition of the disorder; \emph{cf.} \citep{Mott:2020}, Assumptions \ref{ass:associative_memory}-\ref{ass:nn_association} and Remark \ref{rem:unified_stimuli}:

\begin{definition}
Autism is an impairment in which cognitive functions are restricted to discrimination, memorization and prediction of tangible properties of the environment.
\end{definition}

\small
\subsubsection*{Acknowledgments}
The author wants to thank Prof.
\href{https://ece.engineering.arizona.edu/faculty-staff/faculty/jerzy-w-rozenblit}{Jerzy Rozenblit} and Prof. \href{https://neurology.arizona.edu/linda-l-restifo-md-phd}{Linda Restifo} from the University of Arizona in Tucson for numerous and influential discussions, \href{https://profiles.stanford.edu/lynn-koegel}{Prof. Lynn Kern Koegel} from Stanford University for her encouraging opinion about the manuscript, and the Reviewers for their inspiring and valuable comments.
\subsubsection*{Funding}
No funds, grants, or other support was received.
\subsubsection*{Contributions}
Przemysław Śliwiński is the author of the entire manuscript.
\subsubsection*{Competing interests}
The author declares no competing interests.
\subsubsection*{Data availability}
No datasets were generated or analyzed during the current study.
\subsubsection*{Ethics, Consent to Participate, and Consent to Publish declarations}
Not applicable.
\bibliographystyle{abbrvnat}
\bibliography{./IoS-bibliography.bib}

@Article{Bail:2024,
  author   = {Bailey, Drew H. and Jung, Alexander J. and Beltz, Adriene M. and Eronen, Markus I. and Gische, Christian and Hamaker, Ellen L. and others},
  title    = {Causal inference on human behaviour},
  issue    = {8},
  pages    = {1448-1459},
  url      = {https://doi.org/10.1038/s41562-024-01939-z},
  volume   = {8},
  journal  = {Nature Human Behaviour},
  year     = {2024},
}

@Article{Jour:2023,
  author   = {Jourdon, Alexandre and Wu, Feinan and Mariani, Jessica and Capauto, Davide and Norton, Scott and Tomasini, Livia and others},
  title    = {Modeling idiopathic autism in forebrain organoids reveals an imbalance of excitatory cortical neuron subtypes during early neurogenesis},
  issue    = {9},
  pages    = {1505-1515},
  url      = {https://doi.org/10.1038/s41593-023-01399-0},
  volume   = {26},
  journal  = {Nature Neuroscience},
  year     = {2023},
}

@Article{Chap:2021,
  author    = {Chapman, Robert and Veit, Walter},
  title     = {The essence of autism: fact or artefact?},
  number    = {5},
  pages     = {1440--1441},
  url       = {https://doi.org/10.1038/s41380-020-00959-1},
  volume    = {26},
  journal   = {Molecular Psychiatry},
  publisher = {Nature Publishing Group},
  year      = {2021},
}

@Article{Kim:2023,
  author   = {Kim, Eun Joo and Kim, Jeansok J.},
  title    = {Neurocognitive effects of stress: a metaparadigm perspective},
  issue    = {7},
  pages    = {2750 - 2763},
  url      = {https://doi.org/10.1038/s41380-023-01986-4},
  volume   = {28},
  journal  = {Molecular Psychiatry},
  year     = {2023},
}

@Article{Lyu:2024,
  author   = {Lyu, Tian-Jie and Ma, Ji and Zhang, Xi-Yin and Xie, Guo-Guang and Liu, Cheng and Du, Juan and others},
  title    = {Deficiency of {FRMD5} results in neurodevelopmental dysfunction and autistic-like behavior in mice},
  issue    = {5},
  pages    = {1253-1264},
  url      = {https://doi.org/10.1038/s41380-024-02407-w},
  volume   = {29},
  journal  = {Molecular Psychiatry},
  year     = {2024},
}

@Article{Dill:2023,
  author   = {Danica Dillion and Niket Tandon and Yuling Gu and Kurt Gray},
  title    = {Can {AI} language models replace human participants?},
  issn     = {1364-6613},
  number   = {7},
  pages    = {597-600},
  url      = {https://doi.org/10.1016/j.tics.2023.04.008},
  volume   = {27},
  journal  = {Trends in Cognitive Sciences},
  year     = {2023},
}

@Article{Kamp:2024,
  author   = {Kamp-Becker, Inge},
  title    = {Autism spectrum disorder in {ICD}-11—a critical reflection of its possible impact on clinical practice and research},
  pages    = {633–638},
  url      = {https://doi.org/10.1038/s41380-023-02354-y},
  volume   = {29},
  journal  = {Molecular Psychiatry},
  year     = {2024},
}

@Article{Lai:2023,
  author   = {Lai, Meng-Chuan},
  title    = {Mental health challenges faced by autistic people},
  issue    = {10},
  pages    = {1620-1637},
  url      = {https://doi.org/10.1038/s41562-023-01718-2},
  volume   = {7},
  journal  = {Nature Human Behaviour},
  year     = {2023},
}

@Article{Bett:2015,
  author    = {Betty P. V. Ho, Jennifer Stephenson and Mark Carter},
  title     = {Cognitive–behavioural approach for children with autism spectrum disorder: A literature review},
  eprint    = {https://doi.org/10.3109/13668250.2015.1023181},
  number    = {2},
  pages     = {213-229},
  url       = {https://doi.org/10.3109/13668250.2015.1023181},
  volume    = {40},
  journal   = {Journal of Intellectual \& Developmental Disability},
  publisher = {Taylor & Francis},
  year      = {2015},
}

@Article{Ekro:1993,
  author  = {Ekroot, L. and Cover, T. M.},
  title   = {The entropy of {M}arkov trajectories},
  doi     = {10.1109/18.243461},
  number  = {4},
  pages   = {1418-1421},
  url     = {https://doi.org/10.1109/18.243461},
  volume  = {39},
  journal = {IEEE Transactions on Information Theory},
  year    = {1993},
}

@Article{Palm:2017,
  author    = {Palmer, Colin J and Lawson, Rebecca P and Hohwy, Jakob},
  title     = {Bayesian approaches to autism: Towards volatility, action, and behavior.},
  number    = {5},
  pages     = {521–542},
  url       = {https://doi.org/10.1037/bul0000097},
  volume    = {143},
  journal   = {Psychological Bulletin},
  publisher = {American Psychological Association},
  year      = {2017},
}

@Article{DaSi:2025,
  author         = {Da Silva, Sergio and Fiebig, Maria and Matsushita, Raul},
  title          = {Opportunity Costs, Cognitive Biases, and Autism},
  issn           = {2392-7674},
  number         = {1},
  url            = {https://doi.org/10.3390/jmms12010011},
  volume         = {12},
  article-number = {11},
  journal        = {Journal of Mind and Medical Sciences},
  year           = {2025},
}

@Article{Pero:2023,
  author   = {Perochon, Sam and Di Martino, J. Matias and Carpenter, Kimberly L. H. and Compton, Scott and Davis, Naomi and Eichner, Brian and others},
  title    = {Early detection of autism using digital behavioral phenotyping},
  issue    = {10},
  pages    = {2489-2497},
  url      = {https://doi.org/10.1038/s41591-023-02574-3},
  volume   = {29},
  journal  = {Nature Medicine},
  year     = {2023},
}

@Article{Noel:2025,
  author   = {Noel, Jean-Paul and Balzani, Edoardo and Acerbi, Luigi and Benson, Julius and Angelaki, Dora and others},
  title    = {A common computational and neural anomaly across mouse models of autism},
  issue    = {7},
  pages    = {1519-1532},
  url      = {https://doi.org/10.1038/s41593-025-01965-8},
  volume   = {28},
  journal  = {Nature Neuroscience},
  year     = {2025},
}

@Article{Nobr:2022,
  author  = {Nobre, Anna C.},
  title   = {Opening Questions in Visual Working Memory},
  number  = {1},
  pages   = {49-59},
  url     = {https://doi.org/10.1162/jocn\_a\_01920},
  volume  = {35},
  journal = {Journal of Cognitive Neuroscience},
  year    = {2022},
}

@Article{Davi:2015,
  author   = {Greg Davis and Kate Plaisted-Grant},
  title    = {Low endogenous neural noise in autism},
  eprint   = {https://doi.org/10.1177/1362361314552198},
  note     = {PMID: 25248666},
  number   = {3},
  pages    = {351-362},
  url      = {https://doi.org/10.1177/1362361314552198},
  volume   = {19},
  journal  = {Autism},
  year     = {2015},
}

@Article{Assa:2024,
  author   = {Assary, Elham and Oginni, Olakunle A. and Morneau-Vaillancourt, Genevieve and Krebs, Georgina and Peel, Alicia J. and Palaiologou, Elisavet and others},
  title    = {Genetics of environmental sensitivity and its association with variations in emotional problems, autistic traits, and wellbeing},
  issue    = {8},
  pages    = {2438-2446},
  url      = {https://doi.org/10.1038/s41380-024-02508-6},
  volume   = {29},
  journal  = {Molecular Psychiatry},
  year     = {2024},
}

@Article{Saue:2021,
  author  = {Sauer, Ann Katrin and Stanton, Janelle and Hans, Sakshi and Grabrucker, Andreas},
  title   = {Autism spectrum disorders: etiology and pathology},
  pages   = {1-15},
  url     = {https://doi.org/10.36255/exonpublications.autismspectrumdisorders.2021.etiology},
  journal = {Exon Publications},
  year    = {2021},
}

@Article{Laws:2014,
  author   = {Lawson, Rebecca P. and Rees, Geraint and Friston, Karl J.},
  title    = {An aberrant precision account of autism},
  issn     = {1662-5161},
  url      = {https://doi.org/10.3389/fnhum.2014.00302},
  volume   = {8},
  journal  = {Frontiers in Human Neuroscience},
  year     = {2014},
}

@Article{Walk:2023,
  author   = {Walker, Edgar Y. and Pohl, Stephan and Denison, Rachel N. and Barack, David L. and Lee, Jennifer and Block, Ned and others},
  title    = {Studying the neural representations of uncertainty},
  issue    = {11},
  pages    = {1857-1867},
  url      = {https://doi.org/10.1038/s41593-023-01444-y},
  volume   = {26},
  journal  = {Nature Neuroscience},
  year     = {2023},
}

@Article{Wigg:2009,
  author    = {Wiggins, Lisa D and Robins, Diana L and Bakeman, Roger and Adamson, Lauren B},
  title     = {Brief report: sensory abnormalities as distinguishing symptoms of autism spectrum disorders in young children},
  number    = {7},
  pages     = {1087--1091},
  url       = {https://doi.org/10.1007/s10803-009-0711-x},
  volume    = {39},
  journal   = {Journal of Autism and Developmental Disorders},
  publisher = {Springer},
  year      = {2009},
}

@Article{Ulja:2017,
  author   = {Uljarević, Mirko and Richdale, Amanda L. and Evans, David W. and Cai, Ru Ying and Leekam, Susan R.},
  title    = {Interrelationship between insistence on sameness, effortful control and anxiety in adolescents and young adults with autism spectrum disorder ({ASD})},
  issue    = {1},
  pages    = {36},
  url      = {https://doi.org/10.1186/s13229-017-0158-4},
  volume   = {8},
  journal  = {Molecular Autism},
  year     = {2017},
}

@Article{Sing:2023,
  author    = {Singer, Alison and Lutz, Amy and Escher, Jill and Halladay, Alycia},
  title     = {A full semantic toolbox is essential for autism research and practice to thrive},
  number    = {3},
  pages     = {497--501},
  url       = {https://doi.org/10.1002/aur.2876},
  volume    = {16},
  journal   = {Autism Research},
  publisher = {Wiley Online Library},
  year      = {2023},
}

@Article{Rams:2023,
  author    = {Ramstead, Maxwell JD and Sakthivadivel, Dalton AR and Heins, Conor and Koudahl, Magnus and Millidge, Beren and Da Costa, Lancelot and Klein, Brennan and Friston, Karl J},
  title     = {On Bayesian mechanics: a physics of and by beliefs},
  number    = {3},
  pages     = {20220029},
  url       = {https://doi.org/10.1098/rsfs.2022.0029},
  volume    = {13},
  journal   = {Interface Focus},
  publisher = {The Royal Society},
  year      = {2023},
}

@Article{Mott:2020,
  author    = {Mottron, Laurent and Bzdok, Danilo},
  title     = {Autism spectrum heterogeneity: fact or artifact?},
  number    = {12},
  pages     = {3178--3185},
  url       = {https://doi.org/10.1038/s41380-020-0748-y},
  volume    = {25},
  journal   = {Molecular {P}sychiatry},
  publisher = {Nature Publishing Group},
  year      = {2020},
}

@Article{Fris:2010,
  author    = {Friston, Karl},
  title     = {The free-energy principle: a unified brain theory?},
  number    = {2},
  pages     = {127--138},
  url       = {https://doi.org/10.1038/nrn2787},
  volume    = {11},
  journal   = {Nature Reviews Neuroscience},
  publisher = {Nature Publishing Group},
  year      = {2010},
}

@Article{Shir:2024,
  author   = {Shiraishi, Taichi and Katayama, Yuta and Nishiyama, Masaaki and Shoji, Hirotaka and Miyakawa, Tsuyoshi and Mizoo, Taisuke and others},
  title    = {The complex etiology of autism spectrum disorder due to missense mutations of {CHD8}},
  issue    = {7},
  pages    = {2145-2160},
  url      = {https://doi.org/10.1038/s41380-024-02491-y},
  volume   = {29},
  journal  = {Molecular Psychiatry},
  year     = {2024},
}

@Article{Wang:2022,
  author  = {Wang, Zhiyong and Liu, Jingjing and Zhang, Wanqi and Nie, Wei and Liu, Honghai},
  title   = {Diagnosis and Intervention for Children With Autism Spectrum Disorder: A Survey},
  doi     = {10.1109/TCDS.2021.3093040},
  number  = {3},
  pages   = {819-832},
  url     = {https://doi.org/10.1109/TCDS.2021.3093040},
  volume  = {14},
  journal = {IEEE Transactions on Cognitive and Developmental Systems},
  year    = {2022},
}

@Article{Mahe:2025,
  author    = {Mah{\'e}, Oc{\'e}ane and Morel-Kohlmeyer, Shasha and Briend, Fr{\'e}d{\'e}ric and Houy-Durand, Emmanuelle},
  title     = {Predictors of Psychotropic Medication Use Among Autistic Adults},
  pages     = {1--12},
  url       = {https://doi.org/10.1007/s10803-025-06966-x},
  journal   = {Journal of Autism and Developmental Disorders},
  publisher = {Springer},
  year      = {2025},
}

@Book{Mack:2003,
  author    = {MacKay, David J.C.},
  title     = {Information Theory, Inference and Learning Algorithms},
  publisher = {Cambridge University Press},
  url       = {https://www.inference.org.uk/mackay/itila/},
  year      = {2003},
}

@Book{Gali:2016,
  author    = {Galitsky, Boris},
  title     = {Computational Autism},
  publisher = {Springer},
  url       = {https://doi.org/10.1007/978-3-319-39972-0},
  year      = {2016},
}

@Book{Amer:2013,
  author    = {{American Psychiatric Association, DSM-5 Task Force}},
  title     = {Diagnostic and {S}tatistical {M}anual of {M}ental {D}isorders},
  edition   = {5th},
  publisher = {American Psychiatric Publishing, Inc.},
  url       = {https://doi.org/10.1176/appi.books.9780890425596},
  year      = {2013},
}

@Book{Back:2014,
  author    = {B{\"a}ck, Allan},
  title     = {Aristotle's Theory of Abstraction},
  publisher = {Springer},
  url       = {https://doi.org/10.1007/978-3-319-04759-1},
  volume    = {73},
  year      = {2014},
}

@Book{Volk:2021,
  author    = {Volkmar, Fred R.},
  title     = {Encyclopedia of Autism Spectrum Disorders},
  publisher = {Springer},
  url       = {https://doi.org/10.1007/978-3-319-91280-6},
  year      = {2021},
}
\end{document}